\documentclass[preprint,12pt,authoryear]{elsarticle}

\usepackage{amssymb}

\usepackage{hyperref}
\usepackage{multirow}
\usepackage{booktabs}
\usepackage{amsmath,amsfonts,amssymb}
\usepackage{algorithm,algorithmic}
\usepackage{color, soul}

\journal{Journal of Systems and Software}

\begin{document}

\begin{frontmatter}



\title{Black-box Adversarial Sample Generation Based on Differential Evolution}


\author[1]{Junyu Lin}
\author[1]{Lei Xu\corref{cor1}}\ead{xlei@nju.edu.cn}
\author[2]{Yingqi Liu}
\author[2]{Xiangyu Zhang}

\address[1]{Department of Computer Science and Technology, Nanjing University, Nanjing, China}
\address[2]{Department of Computer Science, Purdue University, West Lafayette, USA}
\cortext[cor1]{Corresponding author}

\begin{abstract}
Deep Neural Networks (DNNs) are being used in various daily tasks such as object detection, speech processing, and machine translation. However, it is known that DNNs suffer from robustness problems --- perturbed inputs called adversarial samples leading to misbehaviors of DNNs.
In this paper, we propose a black-box technique called {\it Black-box Momentum Iterative Fast Gradient Sign Method} (BMI-FGSM) to test the robustness of DNN models. The technique does not require any knowledge of the structure or weights of the target DNN. Compared to existing white-box testing techniques that require accessing model internal information such as gradients, our technique approximates gradients through {\em Differential Evolution} and uses approximated gradients to construct adversarial samples.
Experimental results show that our technique can achieve 100\% success in generating adversarial samples to trigger misclassification, and over 95\% success in generating samples to trigger misclassification to a specific target output label. It also demonstrates better perturbation distance and better transferability. Compared to the state-of-the-art black-box technique, our technique is more efficient. Furthermore, we conduct testing on the commercial Aliyun API and successfully trigger its misbehavior within a limited number of queries, demonstrating the feasibility of real-world black-box attack.
\end{abstract}




\end{frontmatter}


%
%
%
\section{Introduction}\label{sec:introduction}

In the past few years, Deep Neural Networks (DNNs)~\citep{lecun2015deep} have achieved great success in many important applications, such as image classification~\citep{krizhevsky2012imagenet}, speech processing~\citep{sak2015fast} and machine translation~\citep{bahdanau2014neural}. Their efficacy even outperforms humans. Modern software applications increasingly include DNNs as a critical component, exampled by autonomous software~\citep{bojarski2016end}, Apple face ID~\footnote{https://machinelearning.apple.com/2017/11/16/face-detection.html}, and Amazon Echo~\footnote{https://developer.amazon.com/alexa/science}. It can be envisioned that DNN model engineering will become an essential step in the software development lifecycle. As such, testing and debugging DNN models is of importance.

However, researchers have revealed that DNNs have robustness problems. That is, they are vulnerable to adversarial samples, i.e., benign inputs that add small and imperceptible perturbation and cause DNNs to misclassify. Adversarial samples hinder the utilization of DNNs in safety-critical systems, especially those related to computer vision, including face recognition~\citep{sharif2017adversarial}, self-driving vehicles~\citep{evtimov2017robust} and medical analysis~\citep{litjens2017survey}. To DNN based applications, adversarial samples are threats but also a method to test DNN models. Our work falls into efficiently and effectively generating adversarial samples to expose robustness problems of DNNs.

Adversarial sample generation methods~\citep{szegedy2013intriguing, goodfellow2014explaining, carlini2017towards, papernot2017practical, su2019one, chen2017zoo} have two categories: {\em white-box} techniques and {\em black-box} techniques. The former requires access to model internals such as model structure, neuron weight values, and gradients. In contrast, the latter treats the subject model as a black box and does not require access to model internals, but rather just model outputs. Black-box techniques have broader applicability. They can be used to test remote applications powered by underlying DNNs. For example, internet service providers such as Alibaba Cloud~\footnote{https://www.alibabacloud.com/} and Google Cloud~\footnote{https://cloud.google.com/} offer pay-for-use APIs, which hide the internal details from users so that white-box techniques are not applicable. As such,  black-box techniques are necessary.

\begin{figure}[!t]
\centering
\includegraphics[width=0.9\linewidth]{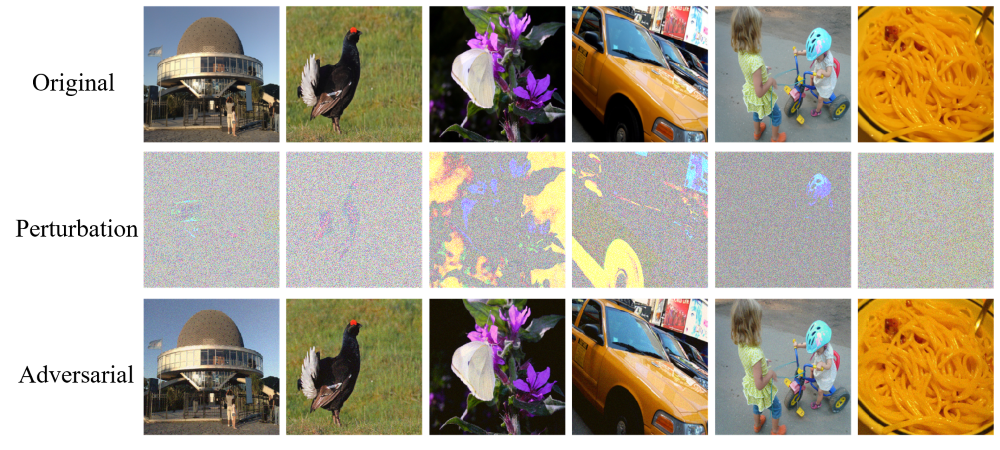}
\caption{Our proposed black-box BMI-FGSM on ImageNet samples. All adversarial samples are misclassified by target DNN (Perturbations are magnified for better visualization).}
\label{fig:rq3_viz}
\end{figure}

In this paper, we develop a new black-box adversarial sample generation method in image classification. For a subject model, we only assume access to model inputs and outputs. Our approach called {\it Black-box Momentum Iterative Fast Gradient Sign Method} (BMI-FGSM) then modifies the original inputs to trigger model misclassification. It aims to achieve the goal with as little mutation as possible, utilizing {\em Differential Evolution}~\citep{storn1997differential} to approximate gradient direction, and leveraging {\it double step size} and {\it candidate reuse} to improve efficiency, which will be explained later.

We compare BMI-FGSM with other state-of-the-art white-box and black-box techniques in both the untargeted setting (i.e., misclassifying to any output label) and the targeted setting (i.e., misclassifying to a specific output label). We also consider the transferability, that is, whether an adversarial sample generated for one model can be used to trigger misbehavior of another model. The experiments on a set of widely used datasets and models show that our proposed BMI-FGSM can achieve a high success rate in both untargeted and targeted settings, comparable or even better than white-box methods. Our approach can generate an adversarial sample within half a minute, faster than other black-box techniques. Finally, we apply our approach to a real-world image recognition system and expose its robustness problem.

In summary, we make the following contributions.
\begin{itemize}
\item We propose a novel black-box adversarial sample generation method named BMI-FGSM. This technique does not require knowledge about model architecture, weight values, or gradients.
\item We propose two novel methods to improve performance, {\it double step size} to enlarge exploration distance and {\it candidate reuse} to approximate momentum that provides guidance for input perturbation, which is critical for generating effective adversarial samples.
\item We compare BMI-FGSM with the state-of-the-art methods. Experimental results on the MNIST, CIFAR10, and ImageNet datasets show that our approach is more efficient than the black-box method ZOO~\citep{chen2017zoo} and has a comparable success rate as the white-box method MI-FGSM~\citep{dong2018boosting}. Furthermore, we use BMI-FGSM to test the commercial Aliyun Image Recognition API and successfully trigger misbehavior.
\end{itemize}

The remainder of this paper is structured as follows. Section~\ref{sec:related} reviews existing research related to adversarial samples. Section~\ref{sec:method} presents our black-box test case generation technique. Section~\ref{sec:experiment} posts three research questions and then answers them with experiments. Section~\ref{sec:conclusion} presents the conclusions and future work.
\section{Related work}\label{sec:related}

Existing works on the DNN robustness problem and adversarial sample generation method in image classification will be reviewed in the following subsections.

\subsection{DNN robustness}
Robustness is critical to the application of DNNs. A popular approach to improving model robustness is to provide additional training and validation data. There are mainly two ways of generating additional data, {\it data augmentation}~\citep{zhong2017random, perez2017effectiveness} and using Generative Adversarial Network (GAN)~\citep{goodfellow2014generative, wang2018high}. The former augments training dataset by transforming original data, e.g., move, rotate, flip, and scale an image to produce new ones. A GAN model is composed of a generator and a discriminator. The generator takes random input and tries to mutate it to a valid input, while the discriminator determines if the mutated one looks like a real input. The two parts compete with each other and ideally the generator would learn to generate real samples. However, existing data augmentation techniques and GANs have limited effectiveness. Hence, a more practical method to gauge and improve DNN robustness is through adversarial samples. Specifically, original input samples are perturbed to generate adversarial samples to trigger model misclassification. The training set can be enhanced with the adversarial samples to retrain the DNN model to improve robustness~\citep{tramer2017ensemble, madry2017towards}. With adversarial training,  DNNs are expected to be less sensitive to noises or perturbations.

\subsection{Adversarial sample generation}

Adversarial sample generation methods can be categorized into white-box methods and black-box methods.

White-box methods assume full knowledge of the target DNN, such as model architecture and neuron weights.
~\citet{szegedy2013intriguing} observed that adding small perturbations to input images can cause DNN model misclassification and converted the generation of adversarial samples to a constrained minimization problem.
~\citet{goodfellow2014explaining} proposed ``fast gradient sign method'' (FGSM), a gradient-based method aiming at a very short generation time.
~\citet{kurakin2016adversarial} proposed a basic iterative method to generate more powerful samples.
~\citet{dong2018boosting} introduced momentum and proposed ``momentum iterative fast gradient sign method'' (MI-FGSM) to further balance the trade-off between success rate and transferability.
~\citet{carlini2017towards} proposed the C\&W method, an optimization-based technique systematically builds examples by directly optimizing the perturbation with an Adam optimizer. Additional proposed mechanisms include binary search and change of variable space.

Black-box methods assume no access to model internals. Instead, they can query the target model by sending inputs and observe the corresponding outputs. Compared with white-box techniques, black-box methods can be used to perform testing to third-party models and have much broader applicability.
~\citet{papernot2017practical} assumed no internal information and insufficient training data, and proposed to train a substitute model with a small synthetic training dataset.
~\citet{narodytska2017simple} discovered the phenomenon that is merely modifying a single pixel might lead to model misclassification.
~\citet{su2019one} exposed DNN robustness problems by leveraging Differential Evolution to search for this kind of pixel.
~\citet{chen2017zoo} proposed Zeroth Order Optimization (ZOO), a derivative-free generation method. The authors exploited a finite differencing method to calculate the approximate gradient by analyzing two very close points in the loss function.

Another type of DNN robustness testing is the transferability testing (or no-box method)~\citep{szegedy2013intriguing, moosavi2017analysis}. These techniques do not query the target model, neither do they generate any samples. Instead, they use adversarial samples produced by another model to test the target DNN. The underlying assumption is that if an adversarial sample can confuse a model, it is likely that it is equally confusing for another model. Therefore, transferability can be considered an important quality metric of generalization for the adversarial test cases.

We aim at developing a black-box adversarial sample generation method that features high success rate, high transferability, and cost-effectiveness.
Many parallel works have also studied the problem of black-box adversarial generation, but our work remains unique in the approach.  
\citet{ilyas2018black} uses a natural evolution strategy (which can be seen as a finite differences estimate on a random gaussian basis) to estimate the gradients for use in the projected gradient descent method.
Following this work, \citet{ilyas2019prior} formalizes the gradient estimation problem and develop a bandit optimization framework incorporating time and data-dependent information, to generate black-box adversarial samples.
In the same threat model, we leverage Differential Evolution to approximate gradient sign to convert a white-box iterative gradient-based method to its black-box version that only requires accessing model outputs.
\citet{alzantot2019genattack} develops a gradient-free approach for generating adversarial examples by leveraging genetic optimization, where the fitness function is defined similarly to CW loss, using prediction scores from the black-box model.
The authors adopt dimensionality reduction and adaptive parameter scaling for boosting gradient-free optimization. In contrast, our approach models the gradient sign, combining with our {\it double step size} and {\it candidate reuse} strategies enables attacks that can reliably generate adversarial samples.
Another line of work aims to generate adversarial samples in different scenarios.  
\citet{cheng2019query} focuses on the label-only setting and propose a generic optimization algorithm, which can be applied to discrete and non-continuous models other than neural networks, such as the decision tree.
\citet{suya2019hybrid} simulates a scenario where the attacker has access to a large pool of seed inputs and proposes a hybrid strategy that combines optimization-based and transferability-based methods. 

\subsection{Testing DNNs}
Compared with traditional software, the behavior of a Deep Learning (DL) system is determined by the structure and weights of DNNs. The dimension and test space of DNN is often larger. DeepXplore~\citep{pei2017deepxplore} proposes a white-box differential testing algorithm for systematically finding inputs that can trigger different behaviors between multiple DNNs. They propose neuron coverage for systematically measuring the parts of a DNN exercised by test inputs. Tensorfuzz~\citep{odena2018tensorfuzz} provides coverage-guided fuzzing methods for neural network by using the approximate nearest neighbor algorithm. DeepTest~\citep{tian2018deeptest} generates test cases that maximize the numbers of activated neurons and finds erroneous behaviors under different realistic driving conditions (e.g., blurring, rain, and fog). DeepGauge~\citep{ma2018deepgauge} proposes multi-granularity testing coverage for DL systems based on the internal state of DNN. Their testing criteria are scalable to complex DNNs. DeepCT~\citep{ma2019deepct} adapts the notion of combinatorial testing and introduces a set of coverage criteria based on neuron input interactions for each layer of DNNs to guide test generation towards balancing the defect detection ability and the number of tests. DeepCover~\citep{sun2018testing} proposes a set of four test criteria for DNNs, inspired by the MC/DC test criteria in traditional software. They evaluate the proposed test criteria on small scale neural networks and show a higher defect detection ability than random testing. DeepMutation~\citep{ma2018deepmutation} adapts the concept of mutation testing and proposes a mutation testing framework specialized for DNNs to measure the quality of test data. They believe that mutation testing is a promising technique that helps to generate higher quality test data.

Existing works on testing the DL system detect erroneous behaviors of DNNs under realistic circumstances  (e.g., lighting, occlusion), and mostly focus on testing criteria, which requires DNNs to be transparent. In contrast, adversarial sample generation demonstrates a particular type of erroneous behavior of DNNs. Our proposed method tests black-box DL systems with adversarial samples that are indistinguishable from the original ones, to expose misbehaviors from the attacker's perspective.
\section{BMI-FGSM algorithm}\label{sec:method}

\begin{figure}[!t]
\centering
\includegraphics[width=1.0\linewidth]{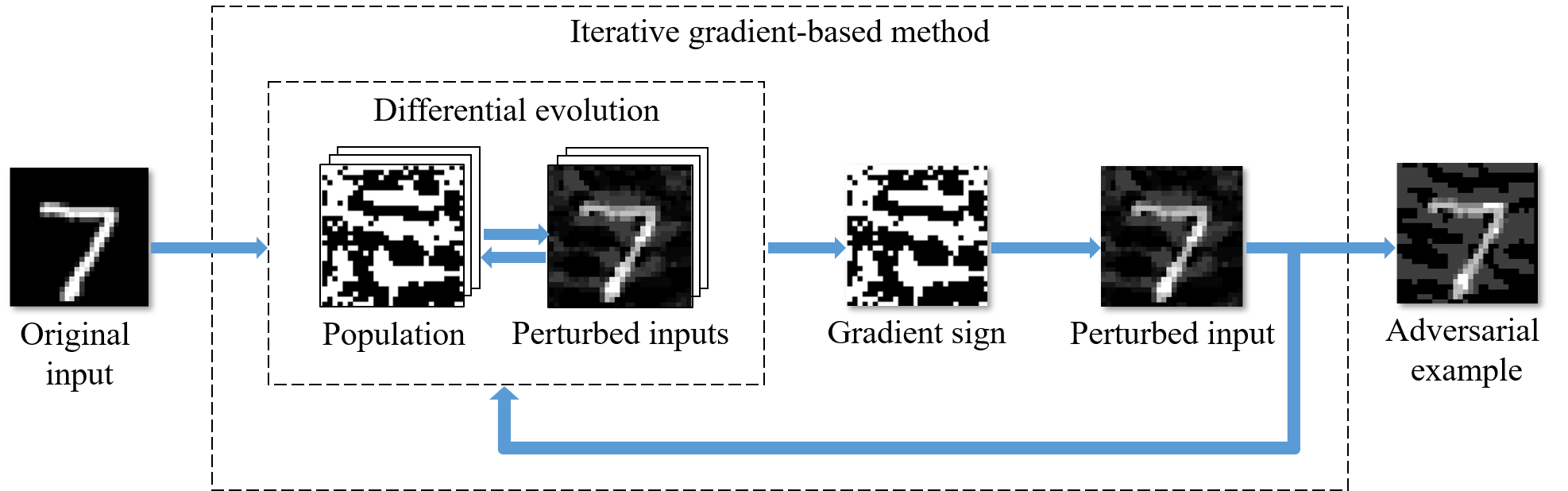}
\caption{Basic framework of the Black-box Momentum Iterative Fast Gradient Sign Method.}
\label{fig:framework}
\end{figure}

Our technique, called {\it Black-box Momentum Iterative Fast Gradient Sign Method} (BMI-FGSM), is inspired by iterative gradient-based methods and {\em Differential Evolution}. It features mechanisms to improve the efficiency and effectiveness of generation. 
Figure~\ref{fig:framework} is the basic framework of our method. Given a clean input image and a pre-trained DNN, Differential Evolution first derives a gradient sign population, children candidates compete with their parents using the corresponding perturbed inputs. Then, we perturb the inputs with the approximate gradient signs. The process repeats until an adversarial sample very similar to the benign input is produced. 

In the following sections, iterative gradient-based methods are first introduced, and Differential Evolution is presented, followed by the complete algorithm with two improvement mechanisms {\it double step size} and {\it candidate reuse}.

\subsection{Iterative gradient-based methods}

Throughout this paper, we focus on a $n$-classifier DNN model~\citep{lecun1998gradient} of image classification task where $I \in \mathbb{R}^m$ is the input image and $F(I) \in \mathbb{R}^n$ is the $n$-dimension output that denotes the predicted labels. More formally, we define a DNN as follows

\begin{equation}
F(I) = \text{softmax}(Z(I))
\end{equation}
\begin{equation}
C(I)=\text{argmax}_i\{F_i(I)\}
\end{equation}

\noindent Here, $Z(I)$ is computed by hidden layers and a set of neuron weight matrices, known as the {\it logits}. The softmax function normalizes the output so $F(I)$ denotes the probability distribution of the predicted labels i.e. $F_i(I)\in[0,1] \wedge \sum_{i}{F_i(I)}=1$, where $F_i(I)$ is the probability that input $I$ belongs to label $i$. Thus the final classification label $C(I)$ is the label with the highest probability.

During the black-box adversarial sample generation, we assume the DNN model is pre-trained and fixed (i.e., neuron weight values are fixed). We assume access to the input image, output label, and output confidence, which is common for black-box methods. We do not assume any access to the structure, neuron weights, or intermediate outputs. Given a clean input $I$ and its ground-truth label $y$, we call $I_{adv}$ an adversarial sample if

\begin{equation}
C(I_{adv}) \ne y \wedge \left\| I_{adv} - I \right\|_p \le \epsilon \quad \text{(untargeted)}
\end{equation}
\begin{equation}
C(I_{adv}) = q \wedge q \ne y \wedge \left\| I_{adv} - I \right\|_p \le \epsilon \quad \text{(targeted)}
\end{equation}

\noindent where in untargeted setting sample is misclassified to any false label and in targeted setting sample misclassified to a specific false label $q$. A threshold $\epsilon$ controls the perturbation magnitude by limiting the $L_p$-norm distance.

Iterative gradient-based methods are white-box generation methods that iteratively calculate the gradient and perturb the input. These methods feature high efficiency, success rate, and extensibility, usually require a small number of back propagations to craft adversarial samples in the $L_\infty$ neighborhood of the original input. For example, the {\it basic iterative method}~\citep{kurakin2016adversarial} produces adversarial sample by applying gradient $T$ times with a small step size:

\begin{equation}\label{eq:bim}
I_t = I_{t-1} + \frac{\epsilon}{T} \cdot \text{sign}(\nabla _{I_{t-1}} J(F(I_{t-1}),y))
\end{equation}

\noindent where $\nabla$ represents the gradient and $J(\cdot)$ is the cost function. It can finish the generation fast and yield superior results, also known as the ``iterative fast gradient sign method''. To further balance the trade-off between success rate and transferability, an improved version of the basic iterative method introduces momentum~\citep{dong2018boosting} by replacing the raw gradient with an accumulated gradient $g_t$:

\begin{equation}\label{eq:accumulate}
g_t = \mu \cdot g_{t-1} + \frac{\nabla _{I_{t-1}} J(F(I_{t-1}),y)}{\left\| \nabla _{I_{t-1}} J(F(I_{t-1}),y) \right\|_1}
\end{equation}

The critical step of iterative gradient-based methods is the gradient computation. In white-box attacks, gradients are usually computed by back propagation.

\subsection{Differential Evolution}

Differential Evolution (DE)~\citep{storn1997differential} is an evolutionary algorithm to solve optimization problems. DE tries to find the input that best fits the objective through input mutation. It has three phases: evolution, crossover, and selection. During the phases of evolution and crossover, new children candidate solutions are generated. In selection, by comparing children with their parents, the ones with better fitness survive and are chosen for the next iteration of evolution. DE has the following advantages: 
(i) DE algorithm is independent of DNN, and hence any improvements of DE~\citep{das2011differential, das2009differential, zhang2009jade} are also directly applicable.
(ii) DE solves non-differentiable, noisy, and dynamic problems such that it can handle both continuous and discontinuous problems without requiring understanding internal structure of the problems.
(iii) DE has a heuristic-based diversity insurance mechanism which can prevent from being trapped in local maxima and minima while standard greedy search and gradient descent often cannot.

\begin{figure}[!t]
\centering
\includegraphics[width=0.9\linewidth]{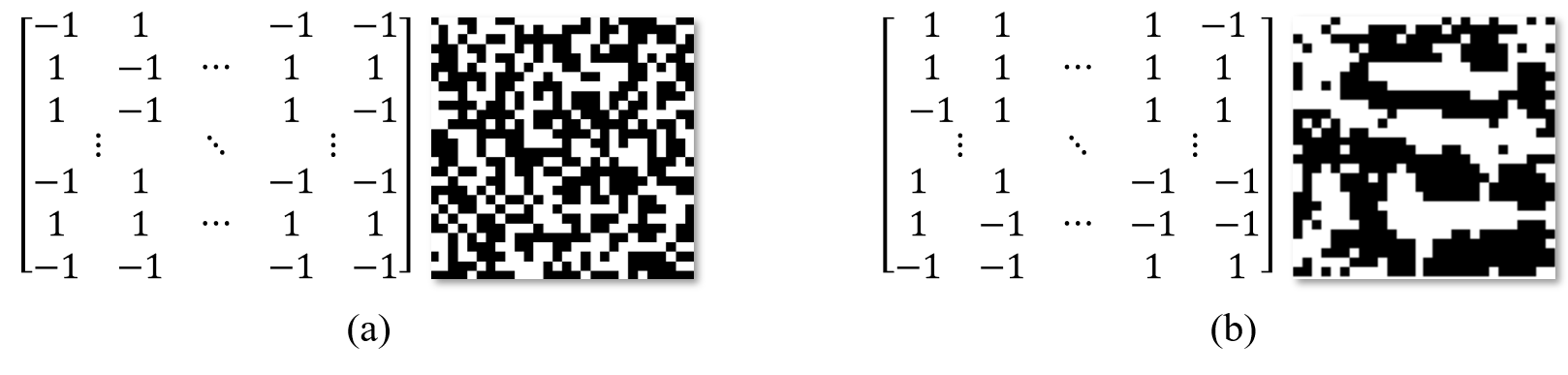}
\caption{Visualization of two candidates in Differential Evolution. White pixels denote the corresponding matrix elements are 1. (a) a candidate in the initial population $x(0)$; (b) a candidate in the last population $x(G)$.}
\label{fig:cand-viz}
\end{figure}

In our algorithm, we employ DE to search for the gradient sign. Denote $I\in\mathbb{R}^m$ as the input image, $y$ as the ground-truth label, $F(I)\in\mathbb{R}^n$ as the outputs of DNN. We initialize the first generation $x(0)$ as follows.

\begin{equation} \label{eq:init}
x(0) = \{x_i(0) | x_{ij}(0)=\text{random}\{-1,1\}; 1\le i\le N; 1\le j\le m\}
\end{equation}

\noindent where $x_i(0)$ denotes the gradient signs with the same size as $I$, and is the $i$-th candidate of the 0-th generation. Variable $x_{ij}(0)$ is the $j$-th feature randomly initialized with 1 or -1. $N$ is the size of population (i.e, the number of candidates in a generation). Figure~\ref{fig:cand-viz}(a) visualizes a candidate in $x(0)$. The candidate is essentially a matrix of the same size as the input image.

The evolution and crossover at the $g$-th generation is defined as follows.

\begin{equation} \label{eq:evolve}
v_i(g+1) = \text{sign}(x_{r_1}(g) + DR \cdot (x_{r_2}(g) - x_{r_3}(g))) \quad (i\ne r_1\ne r_2\ne r_3)
\end{equation}

\begin{equation} \label{eq:crossover}
u_{ij}(g+1) =
\begin{cases}
v_{ij}(g+1) & \text{rand}(0,1)<CR\\ 
x_{ij}(g) & \text{otherwise} 
\end{cases}
\end{equation}

\noindent where $DR$ denotes the scaling factor of differential vector, $CR\in[0,1]$ denotes the crossover probability to control crossover variants. Intuitively, Eq.~\ref{eq:evolve} denotes that a mutant $v_i(g+1)$ is derived from three randomly selected candidates from the previous generaion $x(g)$, denoted by the three random numbers $r_1$, $r_2$, and $r_3$. We assign a sign function because it returns only two valid values representing two different gradient directions, and leads to fast convergence. Eq.~\ref{eq:crossover} denotes that an offspring $u_i(g+1)$ is formed by recombining the mutant with the previous candidate, and $u_{ij}(g+1)$ is the $j$-th feature of $u_i(g+1)$.

Next, compare fitness of $x(g)$ and $u(g+1)$ to determine the next generation $x(g+1)$. However, it is hard to set a fitness function for candidates since they represent gradient signs, so we convert candidates to a set of perturbed images first (Eq.~\ref{eq:bim}, replace $\text{sign}(\cdot)$ with candidate), and then apply the fitness function. For the DNNs we test, although making the confidence of $y$ as fitness works acceptably well, we ultimately settle on this following fitness function.

\begin{equation} \label{eq:fitness}
f(I') = 
\begin{cases}
F_y(I') - \max_{i \ne y}\{F_i(I')\} & \text{untargeted}\\ 
\max_{i \ne q}\{F_i(I')\} - F_q(I') & \text{targeted}
\end{cases}
\end{equation}

\noindent The lower the $f(I')$, the higher fitness value an input $I'$ has. This fitness function aims to suppress the probability of the ground-truth label while enhancing the maximum probability among other false labels. Figure~\ref{fig:cand-viz}(b) visualizes one of the candidates which have evolved $G$ generations. We can observe the pattern that serves as approximate gradient signs.

\begin{algorithm}[t]
\renewcommand{\algorithmicrequire}{\textbf{Input:}}
\renewcommand{\algorithmicensure}{\textbf{Output:}}
\footnotesize
\caption{Approximate gradient signs}
\label{alg:1}
\begin{algorithmic}[1]
    \REQUIRE base sample $I$, generation $x(0)$, perturbation distance $\alpha$, DE iterations $G$;
    \ENSURE last generation $x(G)$;
    \STATE $I(0) \Leftarrow I + \alpha \cdot x(0)$;
    \FOR {$g = 0 : G-1$}
        \STATE Generate next generation $u(g+1)$ from $x(g)$;
        \STATE $I(g+1) \Leftarrow I + \alpha \cdot u(g+1)$;
        \STATE Compare $f(I(g))$ with $f(I(g+1))$ and select $x(g+1)$;
    \ENDFOR
    \RETURN last generation $x(G)$;
\end{algorithmic}
\end{algorithm}

\subsection{Black-box Momentum Iterative Fast Gradient Sign Method} \label{sec:bmifgsm}

Using DE, we leverage Algorithm~\ref{alg:1} to approximate the gradient signs. Children candidates are generated by Eq.~\ref{eq:evolve} and Eq.~\ref{eq:crossover}. We cannot directly compare candidates, so we convert candidates to perturbed images (line 4). The perturbed images' fitness values are compared to determine the offspring using Eq.~\ref{eq:fitness}. Candidate with the lowest fitness score of $x(G)$ is the final approximate gradient signs. Note that in black-box testing, the additional overhead by DE is inevitable because in white-box testing the ``exact gradient signs'' can be computed by back propagation in milliseconds. 
Note that in the black-box attack context, precise estimation of gradient signs is neither possible nor necessary.

\begin{algorithm}[t]
\renewcommand{\algorithmicrequire}{\textbf{Input:}}
\renewcommand{\algorithmicensure}{\textbf{Output:}}
\footnotesize
\caption{BMI-FGSM}
\label{alg:2}
\begin{algorithmic}[1]
\REQUIRE clean example $I_0$, population size $N$, perturbation distance $\epsilon$, DE iterations $G$, iterative gradient-based method iterations $T$, candidate keeping rate $KR$;
    \ENSURE adversarial sample $I_T$;
    \STATE Initialize $N$ candidates $x(0)$\;
    \STATE $\alpha \Leftarrow \epsilon$\;
    \STATE $\beta \Leftarrow \epsilon / T$\;
    \FOR {$t = 0 : T-1$}
        \STATE $x(G) \Leftarrow ApproxGradientSigns(I_t, x(0), \alpha, G)$;
        \STATE $I(G) \Leftarrow I_t + \beta \cdot x(G)$;
        \STATE $I_{t+1} \Leftarrow \underset{i \in I(G)}{\text{argmin}} f(i)$;
        \STATE Keep candidates of best $KR*100\%*N$ fitness in $x(G)$ and initialize other candidates;
        \STATE $x(0) \Leftarrow x(G)$;
        \STATE $\alpha \Leftarrow \alpha - \beta$;
    \ENDFOR
    \RETURN adversarial sample $I_T$;
\end{algorithmic}
\end{algorithm}

We then extend our work to BMI-FGSM, an algorithm inheriting the advantages of both DE and iterative gradient-based method. As shown in Algorithm~\ref{alg:2}, the loop denotes the main procedure of the iterative gradient-based method. After approximating gradient signs with a base sample $I_t$, we apply $x(G)$ to $I_t$ again and generate perturbed images $I(G)$, the next base sample $I_{t+1}$ will be the fittest one of $I(G)$. After $T$ iterations, image $I_T$ is returned as the adversarial sample. The most challenging part of BMI-FGSM is to escape local optimums and re-gain momentum for better results. Hence, we design two mechanisms called {\em double step size} and {\em candidate reuse}.

\subsubsection{Double Step Size}

\begin{figure}[!t]
\centering
\includegraphics[width=0.7\linewidth]{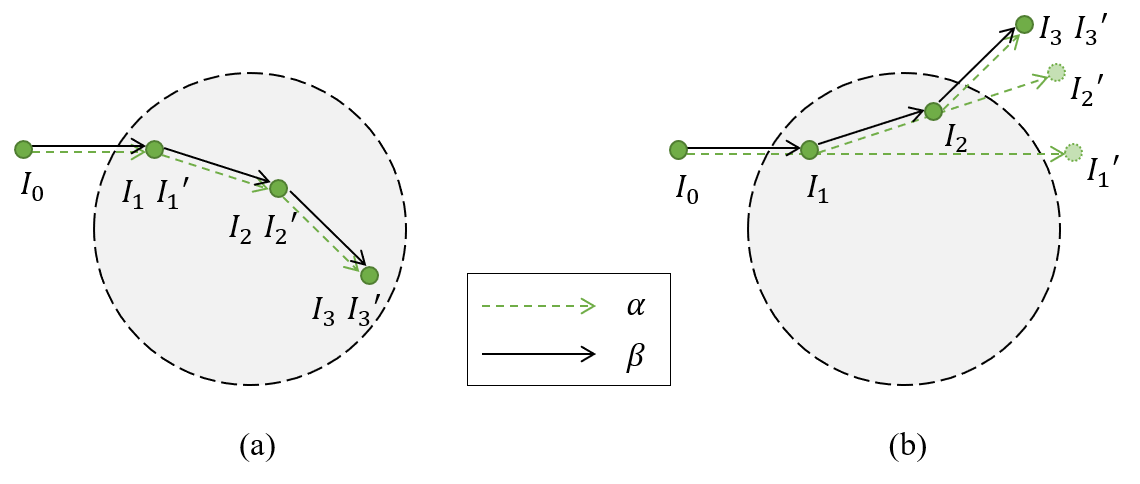}
\caption{An example of double step size. (a) no double step size; (b) double step size.}
\label{fig:double_step}
\end{figure}

There are two types of perturbations in our proposed BMI-FGSM. The first one happens in approximate gradient signs (Algorithm~\ref{alg:1}, line 4). It is a ``temporary perturbation'' in which we always perturb the same base sample $I$, and generate a set of temporary images $I(g+1)$ for fitness evaluation. The other one is in the loop of iterative gradient-based method (Algorithm~\ref{alg:2}, line 6). It is the ``permanent perturbation'' in which we iteratively perturb the clean image from $I_0$ to $I_T$. 

Either perturbation requires a perturbation distance, usually it is $\epsilon / T$, but there would be a problem if we just use the same distance --- the DE loop would evaluate candidates using temporary images that are perturbed with a very small distance. The DE procedure hence runs the risk of being stuck in some local optima. Note that the white-box version of iterative gradient-based method does not suffer from this step size problem as they always calculate exact gradient signs by back propagation.

As a result, we design the {\it double step size}: exploiting two distances for different perturbations. Specifically, $\alpha$ initialized with $\epsilon$ is for gradient signs calculation, and $\beta$ initialized with $\epsilon / T$ is for the permanent perturbation. At the end of iteration $t$, we have perturbed $I_t$ to $I_{t+1}$ with a distance $\beta$, and hence decreased the exploration distance by updating $\alpha \leftarrow \alpha - \beta$ to satisfy $\alpha + t \cdot \beta \equiv \epsilon$, where parameter $\epsilon$ is still the only overall perturbation distance of adversarial sample. Figure~\ref{fig:double_step} shows the difference with and without double step size. The circle area denotes local optima. Points $I_*$ represent the movement of permanent perturbations. Points $I_*'$ denote the best position that DE can find. Without double step size, perturbations are short-sighted so that the path would be trapped for a while. With double step size, we can explore further and hence guide the procedure to escape from the local optima area.

Intuitively, the double step size enables us to use a more considerable perturbation distance when approximating gradient signs. Dynamic exploration distance enlarges the potential to achieve better fitness. With the assistance of double step size, we improve the success rate of sample generation on ImageNet.

\subsubsection{Candidate Reuse}

\begin{figure}[!t]
\centering
\includegraphics[width=0.5\linewidth]{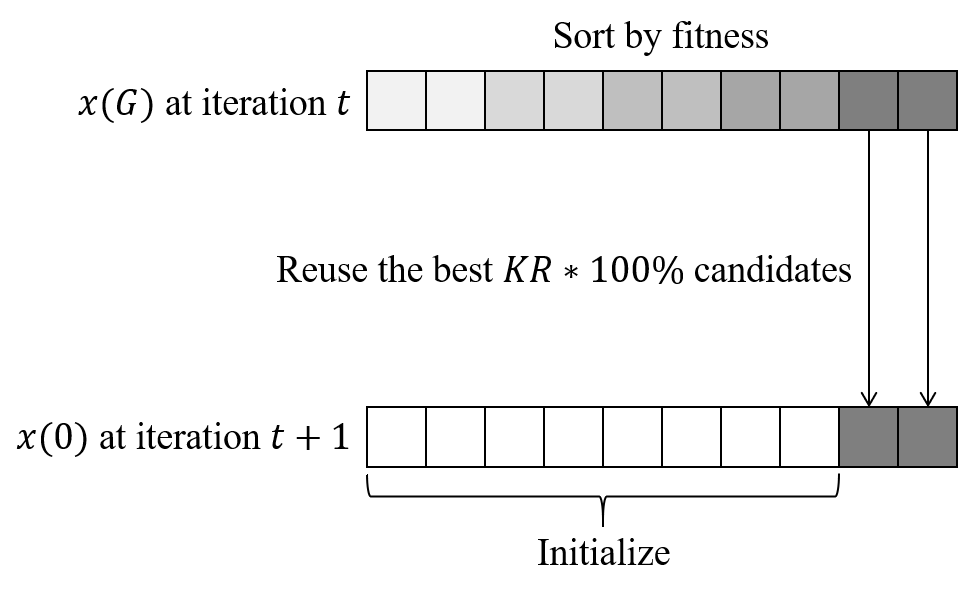}
\caption{An example of candidate reuse.}
\label{fig:candidate_reuse}
\end{figure}

In the training of DNNs, momentum~\citep{polyak1964some} is introduced to improve gradient descent. Using accumulated gradients rather than the raw gradients helps escape local optima and converge fast.  Researchers in the white-box MI-FGSM project~\citep{dong2018boosting} leverage this idea and integrate momentum in their techniques. Accumulated gradient balances the trade-off between transferability and perturbation distance, producing high-quality adversarial samples. But according to our tests, the way of constructing an accumulated gradient cannot be directly ported to our method for two reasons: 
(i) The accumulation of approximate gradient signs also means the accumulation of errors. 
(ii) Candidates have only two valid values 1 and -1, so the accumulated gradient Eq.~\ref{eq:accumulate} is not appropriate.

We design a novel mechanism {\it candidate reuse} to achieve effect similar to gradient accumulation. Specifically, while generation $x(G)$ holds $N$ candidate gradients, our technique saves the best $KR*100\%*N$ candidates as part of the initial population of next round, where $KR$ is the keeping rate. Figure~\ref{fig:candidate_reuse} gives an example of how candidate reuse works. The reused candidates that carry previous iterations' gradient infomation, participate in the evolution and play the role of ``momentum''. The candidate reuse method is more practical in the black-box attack context than the accumulated gradient. Applying candidate reuse, we make BMI-FGSM available on ImageNet.
\section{Experiments}\label{sec:experiment}

We use an Intel Xeon E5 CPU and an Nvidia Tesla K80 GPU for all model trainings and sample generations. We aim to answer the following three research questions:
\begin{itemize}
\item RQ1: How does BMI-FGSM perform compared with existing methods?
\item RQ2: How is the transferability of BMI-FGSM?
\item RQ3: How do double step size and candidate reuse improve BMI-FGSM on large model and dataset?
\item RQ4: How dose BMI-FGSM perform testing third-party applications?
\end{itemize}
We first report that BMI-FGSM is substantially faster than existing black-box techniques and achieves comparable performance with existing white-box methods. Then, we show that the two novel mechanisms do improve performance. Finally, we present the results of generating adversarial samples on a third-party system: the Aliyun Image Recognition API.

\subsection{RQ1: Performance Comparison} \label{sec:rq1}

\subsubsection{Dataset and Models}

For the MNIST~\citep{lecun1998gradient} and CIFAR10~\citep{krizhevsky2009learning} datasets. We randomly choose 100 images for evaluation and sample 10 images from each class. For each image, we apply one untargeted attack and nine targeted generations for the nine other classes. Thus there are 100 untargeted samples and 900 targeted samples in total for each dataset.

The target DNN models we used are based on the settings of Carlini and Wagner (\citet{carlini2017towards}, Table 1). We train and get 99\% accuracy for MNIST and 84\% accuracy for CIFAR10. Data were normalized before classification. Misclassified clean images are ignored during generation.

\subsubsection{Comparison Methods}

On MNIST and CIFAR10, our proposed BMI-FGSM is compared with three state-of-the-art adversarial sample generation techniques C\&W~\citep{carlini2017towards}, ZOO~\citep{chen2017zoo} and MI-FGSM~\citep{dong2018boosting} in both untargeted and targeted settings. For all comparison methods, we refer to the code implementation of Foolbox\footnote{https://github.com/bethgelab/foolbox} with necessary modifications.
\begin{itemize}
\item C\&W (\citet{carlini2017towards}, white-box, optimization-based, $L_2$-norm) builds examples by directly optimizing the perturbation with an optimizer.
\item ZOO (\citet{chen2017zoo}, black-box, optimizaiton-based, $L_2$-norm) exploits finite difference method to calculate approximate gradient.
\item MI-FGSM (\citet{dong2018boosting}, white-box, gradient-based, $L_\infty$-norm) introduces momentum to improve the basic iterative method.
\end{itemize}

\begin{table}[!t]
\footnotesize
\caption{Evaluation of adversarial sample generation methods on MNIST.}
\label{tab:rq1_1}
\tabcolsep 6pt 
\begin{tabular*}{\textwidth}{lrcrrcr}
\toprule
\multirow{2}{*}{Methods} & \multicolumn{3}{c}{Untargeted} & \multicolumn{3}{c}{Targeted}\\
\cmidrule(lr){2-4} \cmidrule(lr){5-7}  
 & Success & Avg. dist & Avg. time & Success & Avg. dist & Avg. time\\
\midrule
C\&W      &100.0\%  &-      &33.4 sec   &100.0\%    &-      &34.0 sec\\
MI-FGSM   &100.0\%  &0.195  & 0.1 sec   &100.0\%    &0.247  &0.2 sec\\
ZOO       &100.0\%  &-      &68.4 sec   &97.8\%     &-      &87.4 sec\\
BMI-FGSM  &100.0\%  &0.227  &16.7 sec   &98.2\%     &0.314  &23.7 sec\\
\bottomrule
\end{tabular*}
\end{table}

\begin{table}[!t]
\footnotesize
\caption{Evaluation of adversarial sample generation methods on CIFAR10.}
\label{tab:rq1_2}
\tabcolsep 6pt 
\begin{tabular*}{\textwidth}{lrcrrcr}
\toprule
\multirow{2}{*}{Methods} & \multicolumn{3}{c}{Untargeted} & \multicolumn{3}{c}{Targeted}\\
\cmidrule(lr){2-4} \cmidrule(lr){5-7}  
 & Success & Avg. dist & Avg. time & Success & Avg. dist & Avg. time\\
\midrule
C\&W      &100.0\%  &-	    &11.3 sec	        &100.0\%    &-	    &12.7 sec\\
MI-FGSM	  &100.0\%  &0.020	&\textless 0.1 sec	&100.0\%	&0.035  &\textless 0.1 sec\\
ZOO	      &100.0\%  &-	    &143.9 sec	        &94.8\%	    &-      &189.4 sec\\
BMI-FGSM  &100.0\%  &0.034	&20.6 sec	        &96.3\%	    &0.047  &30.4 sec\\
\bottomrule
\end{tabular*}
\end{table}

\subsubsection{Parameters}

For C\&W, we conduct 9 binary searches for the best scale parameter $c$ starting from 0.01. We use $\eta=0.01,\beta_1=0.9,\beta_2=0.999$ for the Adam optimizer. The confidence $k=10$ for robustness and the max perturbation distance is set to 20. We run 1000 iterations for both MNIST and CIFAR10. The optimization terminates if the loss does not decrease after each tenth of the iterations.
For ZOO, we use a batch size of 128 i.e., evaluate 128 gradients and update 128 pixels at each step. The ZOO attack is based on C\&W's framework, so we keep the setting consistent with C\&W except running 3000 iterations for MNIST and 1000 iterations for CIFAR10.

For MI-FGSM, we set the maximum iterations $T=40$ for MNIST and $T=20$ for CIFAR10, and use a step size $\epsilon/T=0.01$ per iteration. The upper bounds of perturbation distance are 0.4 and 0.05, respectively. The decay factor $\mu=1$. We check the adversarial sample at the end of each iteration and return early if it successfully causes misclassification. 
For our BMI-FGSM, we set $DR=1, CR=0.9, N=100, KR=0.2$. In the case of MNIST, we set $\epsilon=0.4, T=40, G=50$. In the case of CIFAR10, we set $\epsilon=0.2, T=20, G=100$. Early return is activated.

\begin{figure}[!t]
\centering
\includegraphics[width=0.95\linewidth]{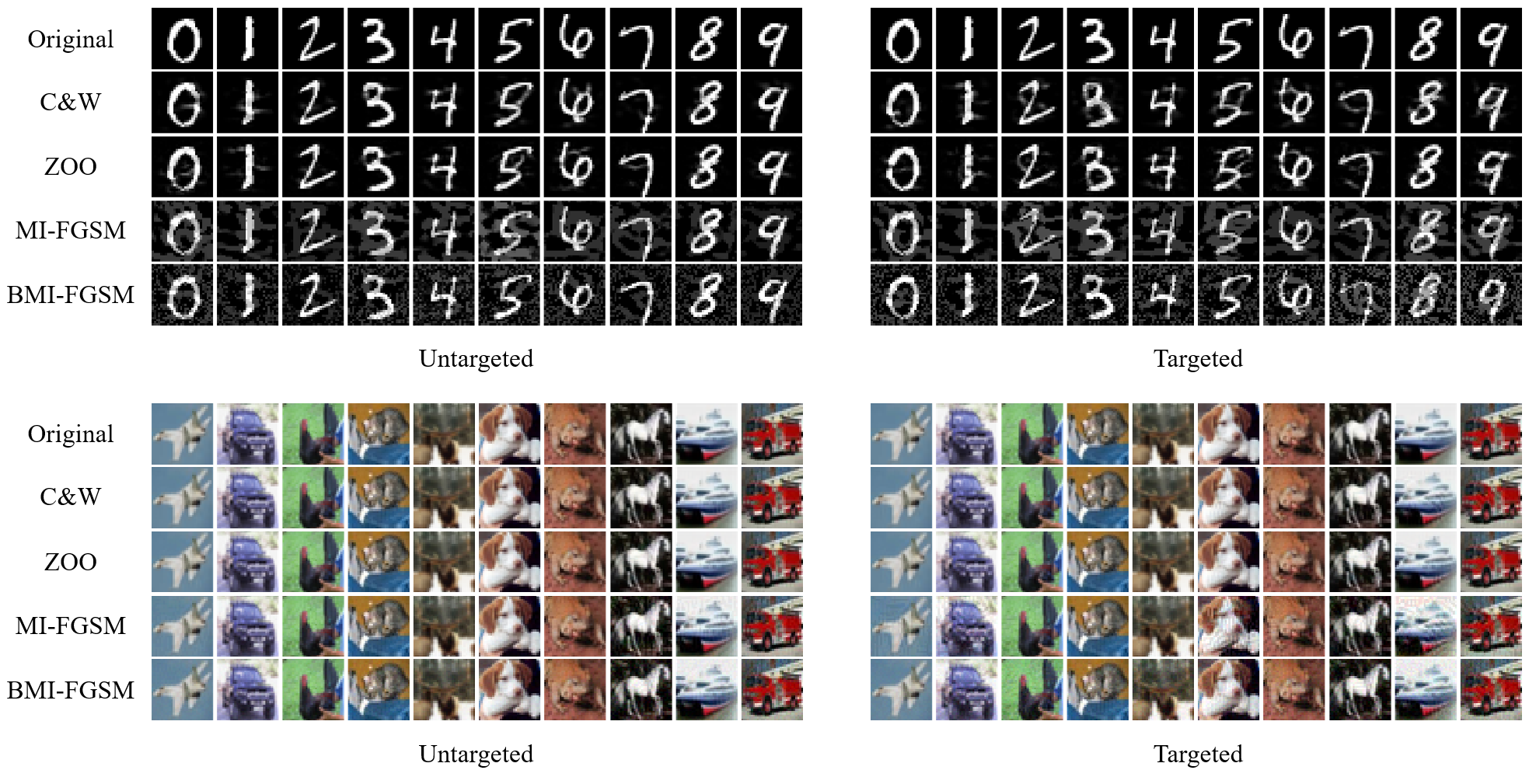}
\caption{Visualization of the adversarial samples on MNIST and CIFAR10.}
\label{fig:rq1_viz}
\end{figure}

\subsubsection{Results}

We report the success rate, average perturbation distance, and time for the four methods on MNIST and CIFAR10 in Table~\ref{tab:rq1_1} and Table~\ref{tab:rq1_2}, respectively.
From column 2, we can observe that all methods achieve 100\% success rates in the untargeted setting. From column 5, The success rate of our approach in the targeted setting is over 95\%, which is higher than the black-box ZOO. 
In terms of the perturbation distance in columns 3 and 6, we ignore the results of C\&W and ZOO because they optimize $L_2$-norm distance while MI-FGSM and BMI-FGSM use $L_\infty$-norm. BMI-FGSM shows very close perturbation distance to the MI-FGSM, that is, our black-box method generates samples with similar quality as the white-box method.
Columns 4 and column 7 illustrate the time cost. Our black-box approach can generates a valid adversarial sample within about 20 seconds for untargeted attacks and 30 seconds for targeted attacks, which is more efficient than the black-box ZOO. Note that our approach is an iterative gradient-based method, which is originally designed for fast sample generation.

Overall, our approach features high success rate and high efficiency. Figure~\ref{fig:rq1_viz} visualizes some generated adversarial samples. We can observe that BMI-FGSM is able to generate images that show a similar visual impression as original images. These malicious samples with imperceptible distortion to human eyes confuse DNNs with high accuracy.

\subsection{RQ2: Transferability} \label{sec:rq2}

\begin{table}[!t]
\footnotesize
\caption{Evaluation of transfer rate on MNIST.}
\label{tab:rq2_1}
\tabcolsep 21pt 
\begin{tabular*}{\textwidth}{lrrrr}
\toprule
\multirow{2}{*}{Methods} & \multicolumn{2}{c}{Untargeted} & \multicolumn{2}{c}{Targeted}\\
\cmidrule(lr){2-3} \cmidrule(lr){4-5}
& FC & LeNet5 & FC & LeNet5 \\
\midrule
C\&W	        &2.0\%	&5.0\%	&6.0\%	&11.0\% \\
MI-FGSM	        &30.0\%	&20.0\%	&42.2\%	&45.4\% \\
ZOO	            &3.0\%	&1.0\%	&3.1\%	&0.2\%  \\
BMI-FGSM	    &38.0\%	&36.0\%	&39.2\%	&24.0\% \\
\bottomrule
\end{tabular*}
\end{table}

\begin{table}[!t]
\footnotesize
\caption{Evaluation of transfer rate on CIFAR10.}
\label{tab:rq2_2}
\tabcolsep 10pt 
\begin{tabular*}{\textwidth}{lrrrrrr}
\toprule
\multirow{2}{*}{Methods} & \multicolumn{3}{c}{Untargeted} & \multicolumn{3}{c}{Targeted}\\
\cmidrule(lr){2-4} \cmidrule(lr){5-7}
 & AllConv & NiN & VGG16 & AllConv & NiN & VGG16\\
\midrule
C\&W	    &12.9\%	&8.2\%	&16.5\% &18.8\%	&11.8\%	&18.8\% \\
MI-FGSM	    &26.8\%	&19.4\%	&31.6\% &46.7\%	&38.3\%	&55.2\% \\
ZOO	        &11.8\%	&9.4\%	&10.6\% &22.1\%	&11.7\%	&14.3\% \\
BMI-FGSM    &21.5\%	&14.5\%	&18.4\% &25.4\%	&22.8\%	&20.1\% \\
\bottomrule
\end{tabular*}
\end{table}

Transferability indicates the cross-model usability of adversarial samples. We evaluate the transferability on MNIST and CIFAR10. We define the transfer rate to be the percentage of transferable adversarial samples generated in Section~\ref{sec:rq1}, i.e., samples that can trigger the misbehavior of another model. 

In the case of MNIST, we train a simple FC network (3x fully connected layers) and a LeNet5~\citep{lecun1998gradient} as targets. In the case of CIFAR10, we train an All Convolution Network (AllConv)~\citep{springenberg2014striving}, a Network in Network (NiN)~\citep{lin2013network} and a VGG16~\citep{simonyan2014very} network as targets. All models above are kept as similar as possible to their original framework with minor modifications at the softmax layer to fit the output.

\subsubsection{Results}

We report the transferability in Table~\ref{tab:rq2_1} and Table~\ref{tab:rq2_2}. Values in the table denote the transfer success rate, represent the generalization of the adversarial samples generated by different methods. We can observe that adversarial samples created by gradient-based methods (MI-FGSM, BMI-FGSM) have a higher transfer rate than those by optimization-based methods (C\&W, ZOO). The optimization-based methods must tolerate larger perturbation distance in order to obtain higher transferability. Besides, our proposed BMI-FGSM takes advantage of momentum, which is introduced for balancing success rate and transferability, achieving a comparable score to white-box MI-FGSM. The transferability of our approach is higher than the black-box ZOO method on both MNIST and CIFAR10.

\subsection{RQ3: Strategies on Large Dataset}\label{sec:rq3}

Modern image classification applications may have a larger dataset and a complex model. Attacks in such settings are challenging and expensive due to the large input space. 
We evaluate the performance and transferability of four methods on the large ImageNet dataset. Then, we study the efficacy of our proposed two mechanisms, double step size and candidate reuse, by generating an image with a hard limit of iterations.

We randomly choose 100 images from ImageNet validation set~\citep{krizhevsky2012imagenet} and apply one untargeted and one targeted generation for each image. Thus there are 100 untargeted samples and 100 targeted samples in total. We use VGG16~\citep{simonyan2014very} as the target model, InceptionV3~\citep{szegedy2016rethinking} and ResNet101~\citep{he2016deep} as the transfer models.
We adapt the CIFAR10 parameter settings in Section~\ref{sec:rq1}. For untargeted setting, an adversarial sample is valid only when its ground-truth label not in the top-5 prediction.

\begin{table}[!t]
\footnotesize
\caption{Adversarial sample generation on ImageNet.}
\label{tab:rq3_1}
\tabcolsep 7pt 
\begin{tabular*}{\textwidth}{lrcrrcr}
\toprule
\multirow{2}{*}{Methods} & \multicolumn{3}{c}{Untargeted} & \multicolumn{3}{c}{Targeted}\\
\cmidrule(lr){2-4} \cmidrule(lr){5-7}  
            & Success   & Avg. dist & Avg. time & Success   & Avg. dist & Avg. time\\
\midrule
C\&W        & 100.0\%	& -         & 8.2 min	        & 98.6\% & -        & 12.0 min \\
MI-FGSM	    & 98.6\%	& 0.003     & \textless 0.1 min	& 98.6\% & 0.005    & 0.1 min \\
ZOO	        & 98.6\%	& -         & 18.1 min	        & 21.6\% & -        & 230.9 min \\
BMI-FGSM	& 98.6\%	& 0.025     & 7.4 min	        & 93.2\% & 0.046    & 19.5 min \\

\bottomrule
\end{tabular*}
\end{table}

\begin{table}[!t]
\footnotesize
\caption{Transfer rate on ImageNet.}
\label{tab:rq3_3}
\tabcolsep 13pt 
\begin{tabular*}{\textwidth}{lrrrr}
\toprule
\multirow{2}{*}{Methods} & \multicolumn{2}{c}{Untargeted} & \multicolumn{2}{c}{Targeted}\\
\cmidrule(lr){2-3} \cmidrule(lr){4-5}
& InceptionV3 & ResNet101 & InceptionV3 & ResNet101 \\
\midrule
C\&W	        &16.2\%	&14.9\%	&15.1\%	&13.7\% \\
MI-FGSM	        &11.0\%	&12.3\%	&11.0\%	&9.6\% \\
ZOO	            &37.0\%	&37.0\%	&25.0\%	&31.3\%  \\
BMI-FGSM	    &28.8\%	&30.1\%	&29.0\%	&33.3\% \\
\bottomrule
\end{tabular*}
\end{table}

\subsubsection{Results}

Table~\ref{tab:rq3_1} demonstrates the performance and Table~\ref{tab:rq3_3} shows the transferability. Our proposed BMI-FGSM achieves a success rate of 98.6\% in the untargeted setting, 93.2\% in targeted setting. Compared with the white-box technique MI-FGSM, our approach has a higher perturbation distance due to large scale input images, and hence more gradients need to be approximated. Compared with the black-box ZOO, our approach significantly reduces time consumption and generates more transferable samples. We plot some of our generated samples in Figure~\ref{fig:rq3_viz}. All are hard to distinguish from the original samples.

The effect of different strategies is illustrated in Table~\ref{tab:rq3_2}. Observe that the average distance and the first valid iteration (the iteration where the first successful adversarial sample was found) have no evident distinction, while the success rates change noticeably -- the success rate decreases by about 28\% without double step size. The algorithm is almost unusable without candidate reuse. We also evaluate an alternative that uses the perturbed images as the population rather than gradient sign, while keeping everything else the same. Observe that the success rate decrease as well, suggesting incompatibility. This result indicates that new strategies are surely needed to apply this alternative.

\begin{table}[!t]
\footnotesize
\caption{Different strategies comparison on ImageNet.}
\label{tab:rq3_2}
\tabcolsep 14pt 
\begin{tabular*}{\textwidth}{lrcc}
\toprule
Strategy    &Success rate   &Avg. dist   & First valid\\
\midrule
BMI-FGSM             & 98.6\%  & 0.025  & 861\\
No double step size  & 70.3\%  & 0.021  & 808\\
No candidate reuse   &    0\%  & -      & -\\
Perturbed images as population & 63.5\% & 0.020 & 797\\
\bottomrule
\end{tabular*}
\end{table}

\begin{figure}[!t]
\centering
\includegraphics[width=0.7\linewidth]{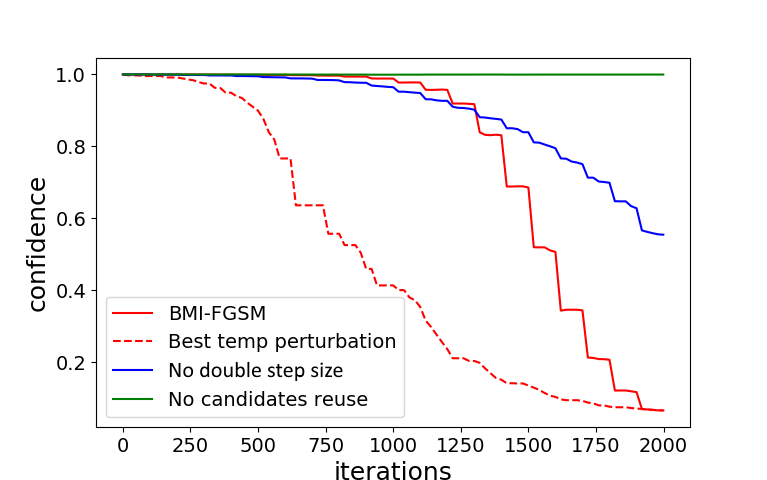}
\caption{Confidence of ground-truth versus iterations.}
\label{fig:rq3_plot}
\end{figure}

To further investigate the advantages of double step size and candidate reuse, we set a hard $T \times G = 2000$ iterations and report the confidence of ground-truth label versus iterations in Figure~\ref{fig:rq3_plot}. The lower ground-truth label confidence, the higher attack ability of sample. The curve of no candidate reuse ends at about 0.99 shows the importance of momentum information. The curve of no double step size ends at about 0.55 because a single short perturbation distance runs the risk of being trapped in some local optima. In contrast, the dashed curve that indicates the best temporary perturbed images when approximate gradient signs drop quickly, and guide BMI-FGSM to achieve lower confidence of 0.05. With all the techniques applied, BMI-FGSM is able to generate visually undetectable adversarial samples that effectively suppress the probability of ground-truth label.

Adversarial samples represent a particular type of misbehavior of the DL system. To illustrate the ability of BMI-FGSM to test the robustness of DNN, we use random testing without coverage guidance and Tensorfuzz~\footnote{https://github.com/brain-research/tensorfuzz} with coverage guidance as baselines of DNN testing techniques. Table~\ref{tab:rq3_4} reports the performance of different methods in generating untargeted adversarial samples. Observe that our approach outperforms random testing and Tensorfuzz, which means BMI-FGSM is more appropriate for robustness testing of adversarial attacks.

\begin{table}[!t]
\footnotesize
\caption{Comparison of testing techniques on ImageNet.}
\label{tab:rq3_4}
\tabcolsep 23pt 
\centering
\begin{tabular*}{0.75\textwidth}{lrc}
\toprule
Strategy    &Success rate   &Avg. dist\\
\midrule
Random               & 0\%     & -\\
Tensorfuzz           & 31.1\%  & 0.046\\
BMI-FGSM             & 98.6\%  & 0.025\\
\bottomrule
\end{tabular*}
\end{table}

\subsection{RQ4: Testing third-party API}

\begin{figure}[!t]
\centering
\includegraphics[width=1.0\linewidth]{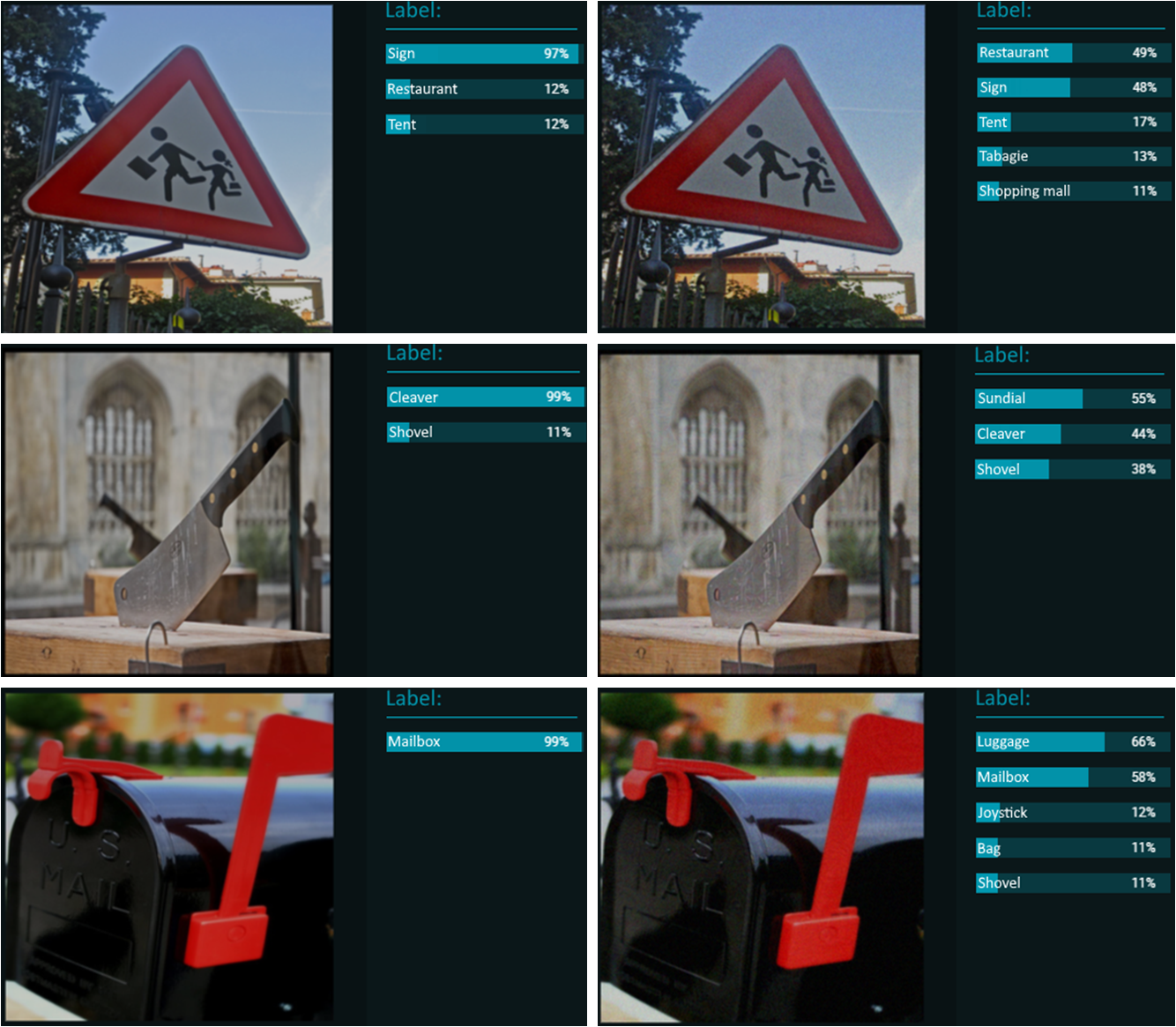}
\caption{Aliyun Image Recognition API. Left: clean image. Right: adversarial image.}
\label{fig:rq4_viz}
\end{figure}

In order to validate the reliability and applicability of our method to third-party applications, we test the Aliyun Image Recognition API~\footnote{https://help.aliyun.com/product/53258.html}, a commercial computer vision toolkit powered by Alibaba Cloud. The API is a n-way classifier that can be queried and outputs label-score pairs for a given image, without any internal details.
Our goal is to perform adversarial sample generation to trigger black-box API misbehavior. However, testing Aliyun API is more challenging than that on a common black-box model because of the following reasons: (i) The Aliyun API provides a free trial of 5000 queries. After that, it costs about \$1.5 per 5000 queries. From the perspective of the attacker,  a query-efficient adversarial generation algorithm is highly desirable. (ii) The number of classes is even larger, but we do not know how many and cannot enumerate all labels. (iii) The API only outputs scores for up to 5 top labels. The scores do not sum to one. They are neither probabilities nor logits.

In this experiment, we adapt the parameter settings in Section~\ref{sec:rq3}. We set the perturbation distance bound to be $\epsilon=0.1$ and perform an untargeted generation with a budget of 5000 queries to the target API. We use API's top-1 prediction of the original image as the fitness. The algorithm early terminates when the top-1 prediction change.

\subsubsection{Results}

Our approach achieves a 92\% success rate against the Aliyun API on an ImageNet sample set of 25 images. Some adversarial samples against the image recognition API are given in Figure~\ref{fig:rq4_viz}. The left part is the original predictions, showing that the API correctly classifies clean images with high confidence. The right part is the adversarial predictions, showing adversarial samples mislabeled by the API. For example, the sign image with a 97\% confidence score is misclassified as a restaurant, while the two images look very similar. Overall, we successfully trigger the Aliyun API misbehavior with perturbed images.
\section{Conclusion}\label{sec:conclusion}
In this paper, we introduce Differential Evolution and develop a new type of black-box adversarial sample generation method called BMI-FGSM. 
We generate adversarial samples that are hard to detect and successfully attack DNNs on MNIST, CIFAR10, and ImageNet. 
We propose two mechanisms, double step size and candidate reuse, to be the essential part of our black-box method and conduct experiments to validate their efficacy. 
Experimental results show that our approach obtains success rate, perturbation distance, and transferability comparable to the state-of-the-art white-box techniques while reducing the time consumption of black-box method considerably. It even achieves similar performance to a white-box method on a large dataset.
Finally, we craft adversarial samples for the Aliyun Image Recognition API and expose its robustness problem, demonstrating that our approach is able to test real-world third-party systems from the perspective of the attacker.

Future work includes testing models with defense techniques or models in other areas. This work also opens up new opportunities on how to combine adversarial techniques with evolutionary algorithms like Differential Evolution.
%
%
%





\end{document}